\documentclass[sigconf]{acmart}

\usepackage{booktabs} 
\newenvironment{myitemize}[1][]{
	\begin{list}{$\bullet$}
		{
			\setlength{\leftmargin}{5mm}
			\setlength{\parsep}{1mm}
			\setlength{\topsep}{0mm}
			\setlength{\itemsep}{0mm}
			\setlength{\labelsep}{1.5mm}
			\setlength{\itemindent}{0mm}
			\setlength{\listparindent}{5mm}
	}}
	{\end{list}}

\usepackage{subfigure}
\usepackage{balance}
\AtBeginDocument{%
  \providecommand\BibTeX{{%
    \normalfont B\kern-0.5em{\scshape i\kern-0.25em b}\kern-0.8em\TeX}}}

\copyrightyear{2021}
\acmYear{2021}
\setcopyright{acmcopyright}\acmConference[MM '21]{Proceedings of the 29th ACM
	International Conference on Multimedia}{October 20--24, 2021}{Virtual Event, China}
\acmBooktitle{Proceedings of the 29th ACM International Conference on Multimedia
	(MM '21), October 20--24, 2021, Virtual Event, China}
\acmPrice{15.00}
\acmDOI{10.1145/3474085.3475289}
\acmISBN{978-1-4503-8651-7/21/10}

\begin{document}
\fancyhead{}
\title{FoodLogoDet-1500: A Dataset for Large-Scale Food Logo Detection via Multi-Scale Feature Decoupling Network}

\author{Qiang Hou}
\affiliation{%
	\institution{School of Information Science and Engineering, Shandong Normal University}
	\streetaddress{No.1 Daxue Road}
	\city{Shandong}
	\country{China}
}
\email{2019309052@stu.sdnu.edu.cn}

\author{Weiqing Min}
\affiliation{%
	\institution{Key Laboratory of Intelligent Information Processing,
		Institute of Computing Technology, Chinese Academy of Sciences}
	\streetaddress{No.6 Kexueyuan South Road}
	\city{Beijing}
	\country{China}}
\email{minweiqing@ict.ac.cn}
\authornotemark[1]

\author{Jing Wang}
\affiliation{%
	\institution{School of Information Science and Engineering, Shandong Normal University}
	\streetaddress{No.1 Daxue Road}
	\city{Shandong}
	\country{China}
}
\email{2018020875@stu.sdnu.edu.cn}

\author{Sujuan Hou}
\affiliation{%
	\institution{School of Information Science and Engineering, Shandong Normal University}
	\streetaddress{No.1 Daxue Road}
	\city{Shandong}
	\country{China}
	\postcode{250358}
}
\email{sujuanhou@sdnu.edu.cn}
\authornote{Corresponding Author.}

\author{Yuanjie Zheng}
\affiliation{%
	\institution{School of Information Science and Engineering, Shandong Normal University}
	\city{Shandong}
	\country{China}
	\postcode{250358}
}
\email{zhengyuanjie@gmail.com}

\author{Shuqiang Jiang}
\orcid{1234-5678-9012-3456}
\affiliation{%
	\institution{Key Laboratory of Intelligent Information Processing,
		Institute of Computing Technology, Chinese Academy of Sciences}
	\streetaddress{No.6 Kexueyuan South Road}
	\city{Beijing}
	\postcode{100190}
	\country{China}
}
\email{sqjiang@ict.ac.cn}

\begin{abstract}
	Food logo detection plays an important role in the multimedia for its wide real-world applications, such as food recommendation of the self-service shop and infringement detection on e-commerce platforms. A large-scale food logo dataset is urgently needed for developing advanced food logo detection algorithms. However, there are no available food logo datasets with food brand information. To support efforts towards food logo detection, we introduce the dataset FoodLogoDet-1500, a new large-scale publicly available food logo dataset, which has 1,500 categories,  about 100,000 images and about 150,000 manually annotated food logo objects. We describe the collection and annotation process of FoodLogoDet-1500, analyze its scale and diversity, and compare it with other logo datasets. To the best of our knowledge, FoodLogoDet-1500 is the first largest publicly available high-quality dataset for food logo detection. The challenge of food logo detection lies in the large-scale categories and similarities between food logo categories. For that, we propose a novel food logo detection method Multi-scale Feature Decoupling Network (MFDNet), which decouples classification and regression into two branches and focuses on the classification branch to solve the problem of distinguishing multiple food logo categories. Specifically, we introduce the feature offset module, which utilizes the deformation-learning for optimal classification offset and can effectively obtain the most representative features of classification in detection. In addition, we adopt a balanced feature pyramid in MFDNet, which pays attention to global information, balances the multi-scale feature maps, and enhances feature extraction capability. Comprehensive experiments on FoodLogoDet-1500 and other two popular benchmark logo datasets demonstrate the effectiveness of the proposed method. The code and FoodLogoDet-1500 can be found at {\color{red}{\url{https://github.com/hq03/FoodLogoDet-1500-Dataset}}}.
\end{abstract}



\maketitle

\section{Introduction}
\begin{figure*}
	\centering
	\includegraphics[width=0.8\textwidth]{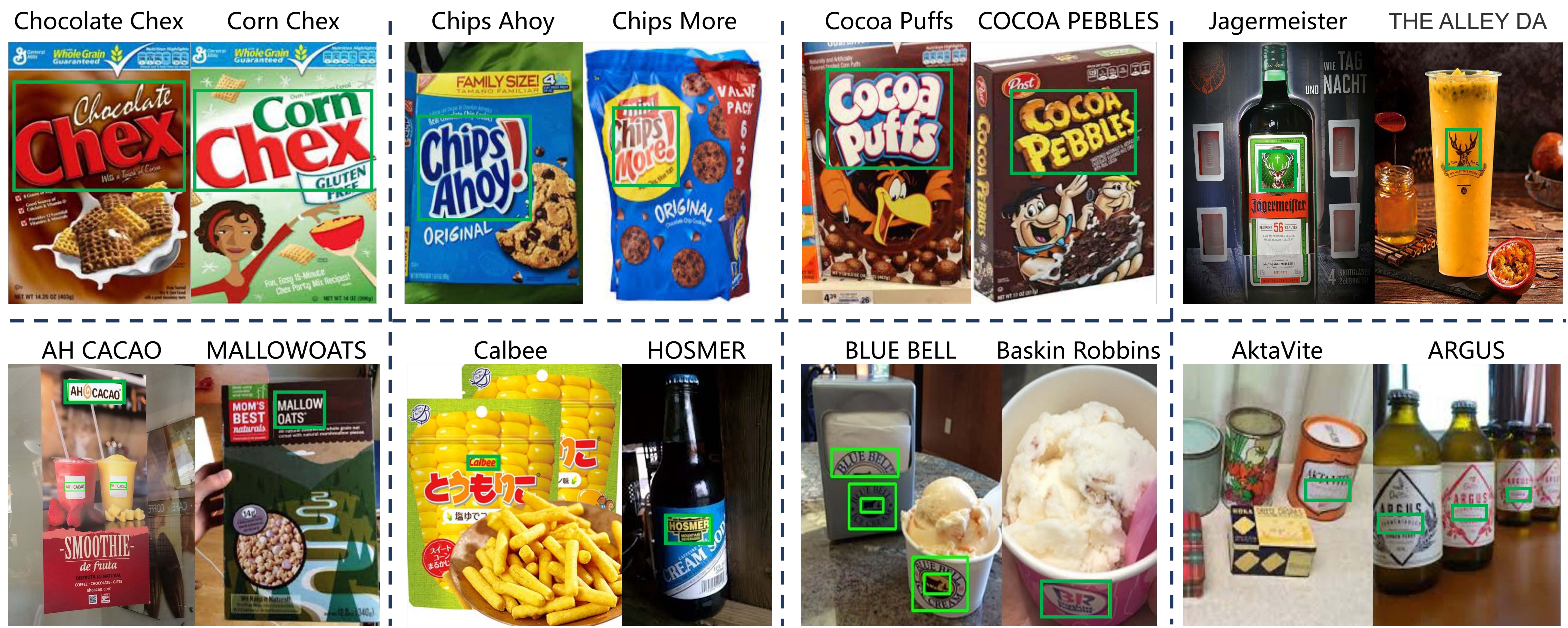}
	\caption{Some samples from FoodLogoDet-1500. Green boxes: ground-truth boxes.}
	\label{challenge}
\end{figure*}
Logo detection research has always been extensively studied in the field of multimedia~\cite{gao2014brand,kalantidis2011scalable,10.1145/2393347.2396358,romberg2013bundle,eggert2015benefit}. As one significant task in logo detection, food logo detection can be applied for healthy diet recommendation, food trademark infringement dispute, food advertising placement and supermarket self-checkout system. For example, with the rapid development of e-commerce platforms, many food businesses ignore copyright awareness in pursuit of profits, resulting in irreparable losses. Through food logo detection, we can avoid trademark infringement by detecting the new food logo and comparing their similarity with existing food logos. Furthermore, when we detect the logo from food products, we can further conduct various health-related and sell advertising applications~\cite{min2019survey}. 

Despite its great potential applications, food logo detection is still a challenging task, and the challenge mainly derives from two aspects:
\begin{myitemize}
	\item \textbf{There is a lack of large-scale food logo dataset for food logo detection.} Existing works mainly focus on messy logo datasets for logo detection, such as FlickrLogos-32~\cite{romberg2011scalable} and QMUL-OpenLogo~\cite{su2018open}. For example, Romberg~\textit{et al.}~\cite{romberg2011scalable} introduce the FlickrLogos-32 dataset with 32 categories but only very few classes belong to food logos. Su~\textit{et al.}~\cite{su2018open} release one logo dataset with 352 categories, and full annotations. However, it is not all about food logos. Existing logo datasets only contain a tiny number of food logo categories. Therefore, they are probably not sufficient to construct more complicated deep learning models for food logo detection.
	\item \textbf{There are multi-scale and similar logos from food logo images, which are harder to detect in many cases.} Compare with other logo images, the multi-scale and similar food logos of the food logo images are more complicated, and make it difficult to accurately extract effective features. As shown in Fig.~\ref{challenge}, the first row represents eight different food logo images. However, two classes of food logos are so similar that they are difficult to distinguish, such as ‘Chips Ahoy’ and ‘Chips More’.  Different brands of the same food may have similar food logos, which makes detection more difficult.
	Some food logos look like text, and they also have occlusion problems. Different food logos also have the characteristics of multi-scale, the second row illustrates this, such as ‘MALLOW OATS’ and ‘Calbee’. These characteristics lead to the difficulty of food logo detection.
	
\end{myitemize}

In this work, we address data limitations by building a large-scale dataset FoodLogoDet-1500 with 1,500 categories, 99,768 images and 145,400 objects. As the largest food logo detection dataset so far, FoodLogoDet-1500 brings great opportunities and challenges for food logo detection in general and sophisticated scenarios. 
To solve another challenge, we propose a Multi-scale Feature Decoupling Network (MFDNet) to improve food logo detection. This is achieved by two main modules named Feature Offset Module (FOM) and Balanced Feature Pyramid (BFP). FOM firstly decouples classification and regression into two branches, and then utilizes the deformation-learning for optimal classification offset. Finally, the optimal classification offset is merged with the original features of the network. The experiment proves that FOM can improve the classification accuracy in the food logo detection. In addition, we adopt BFP in MFDNet, which pays attention to global information and has a good performance on multi-scale food logos.

To summarize, our paper main contributions are as follow:

\begin{myitemize}
	\item We first introduce a large-scale and highly diverse food logo
	dataset FoodLogoDet-1500 with 1,500 categories, 99,768 images and 145,400 objects.
	\item We propose a Multi-scale Feature Decoupling Network for food logo detection by decoupling shared heads of classification and regression at the same time. In this network, we further introduce a balanced feature pyramid to ensure the detection of multi-scale food logos.
	\item We conduct extensive evaluation on three datasets, including FoodLogoDet-1500 and other two standard logo datasets QMUL-OpenLogo, FlickrLogos-32, and verify the effectiveness of our proposed method.
\end{myitemize}
\section{Related Work}
\begin{figure*}
	\centering
	\includegraphics[width=0.8\textwidth]{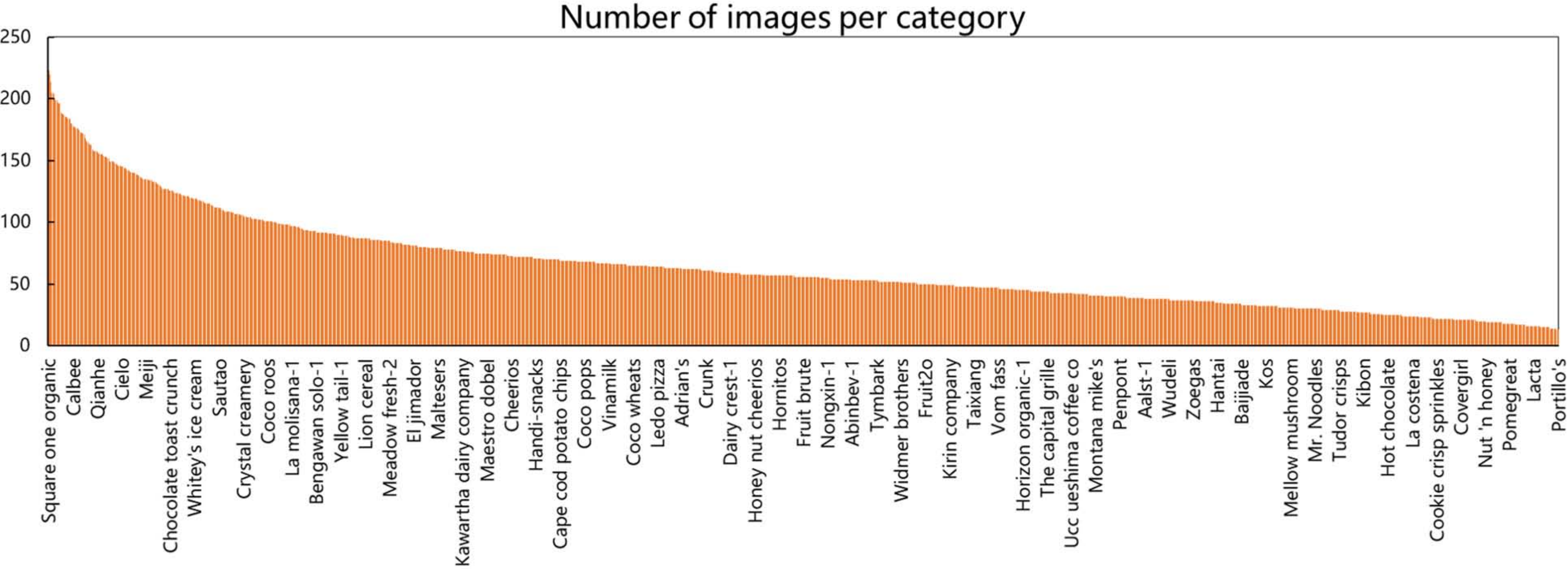}
	\caption{Sorted distribution of the number of images from each food logo in the FoodLogoDet-1500.}
	\label{distribution}
\end{figure*}
\begin{table}[t]	
	\caption{Comparison between FoodLogoDet-1500 and existing logo datasets.}
	\label{datasets_summary}
	\small
	\begin{center}
		\begin{tabular}{c|c|c|c|c}
			\hline
			Dataset&$\sharp$Logos&$\sharp$Images&$\sharp$Objects& Availability\\
			\hline
			\hline
			BelgaLogos~\cite{joly2009logo} &37& 10,000&-&Yes\\
			FlickrLogos-27~\cite{kalantidis2011scalable} &27&1,080 &4,671&Yes\\
			FlickrLogos-32~\cite{romberg2011scalable} &32&8,240&5,644&Yes\\
			Top-Logo-10~\cite{su2017deep}&10&700&-&Yes\\
			WebLogo-2M~\cite{su2017weblogo}&194&1,867,177&-&Yes\\
			QMUL-OpenLogo~\cite{su2018open}&352&27,083&-&Yes\\
			Logos-in-the-Wild~\cite{tuzko2017open}&871&11,054&32,850&Yes\\
			Logo-2K+~\cite{wang2020logo}&2,341&167,140&-&Yes\\
			LogoDet-3K~\cite{wang2020logodet}&3,000&158,652&194,261&Yes\\
			MICC-Logos~\cite{sahbi2012context}&13&720&-&No\\
			FlickrBelgaLogos~\cite{letessier2012scalable} &34&10,000& 2,695&No\\
			Logo-18~\cite{2015LOGO}&18&8,460&16,043&No\\
			Logo-160~\cite{2015LOGO}&160&73,414&130,608&No\\
			Logos-32plus~\cite{bianco2017deep}&32&7,830&12,302&No\\
			Video SportsLogo~\cite{liao2017mutual}&20&2,000&-&No\\
			CarLogo-51~\cite{2014Fast}&51&11,903&-&No\\
			Open Brands~\cite{jin2020open}&1,216&1,437,812&3,113,828&No\\
			SynthLogo~\cite{MD2018Logo}&604&280,000&-&No\\
			PL2K~\cite{fehervari2019scalable}&2,000&295,814&-&No\\
			\hline
			\textbf{FoodLogoDet-1500}&\textbf{1,500}&\textbf{99,768}&\textbf{145,400}&\textbf{Yes}\\
			\hline
		\end{tabular}
	\end{center}
	
\end{table}
This section presents related work in the areas of logo datasets and logo detection.

\subsection{Logo Datasets} The large-scale datasets play an indispensable role in current object detection algorithms, and it is no exception in food logo detection.
In object detection, MS COCO~\cite{lin2014microsoft} and PASCAL VOC~\cite{Everingham2010voc} are the most commonly used datasets. In logo detection, FlickrLogos-32~\cite{romberg2011scalable} is the most popular dataset. However, it only consists of 32 logo categories with 70 images in each category. Similarly, Top-Logo-10~\cite{su2017deep} contains fewer logo categories and fewer images. Logo-2K+~\cite{wang2020logo} belongs to image-level dataset and cannot be used for logo detection. QMUL-OpenLogo~\cite{su2018open} consists of 352 logo categories, but it is an all-encompassing logo dataset (e.g., Foods, Clothes, Necessities), and it can not be used for food logo detection. Wang~\textit{et al.} introduce LogoDet-3K~\cite{wang2020logodet}, which has 3,000 categories, where only 932 classes belong to food logos. In order to promote the development of food logo detection in the multimedia community, we make supplementary improvements on food logos of LogoDet-3K. And then FoodLogoDet-1500 was built completely. In addition, some researchers construct other logo datasets, such as Logo-160~\cite{2015LOGO} and Open Brands~\cite{jin2020open}. These logo datasets 
are not currently available to the public, and are not helpful to the development of logo detection research. 

Food logo detection is an important branch of logo detection. However, those have not publicly available food logo datasets with brand information at present. Therefore, we introduce a new large-scale food logo dataset FoodLogoDet-1500 with 1,500 food logo categories. Table~\ref{datasets_summary} summarizes the statistics of existing logo datasets and FoodLogoDet-1500. To the best of our knowledge, FoodLogoDet-1500 is the first largest publicly available high-quality dataset for food logo detection. It helps to promote the development of food logo detection research.

\subsection{Logo Detection} Typically, logo detection is performed by adapting object detection methods in the domain of commercial logos~\cite{iandola2015deeplogo} (i.e., treating each logo as a different object or class). Traditionally, the use of hand-crafted features, such as SIFT~\cite{lowe1999object} and textures~\cite{haralick1973textural}, along with statistical classifiers, such as Support Vector Machines (SVM)~\cite{cortes1995support}, have been the main approaches for object detection. In the last few years, deep learning has shown its good performance in object detection, and Convolutional Neural Networks (CNNs)~\cite{lecun2015deep,schmidhuber2015deep} was the core element of deep learning methods. Multiple methods for object detection using CNNs have been presented. In general, object detectors could be divided into two types: two-stage detector and one-stage detector. Two-stage means that the object detection algorithm needs to be completed in two steps. First, candidate regions need to be obtained and then classified, such as R-CNN series like Fast R-CNN~\cite{girshick2015fast} and Faster R-CNN~\cite{ren2015faster}. On the other hand, one-stage detector, which can be understood as one-step detection, does not need to search for candidate regions separately, typically including SSD~\cite{liu2016ssd} and YOLO~\cite{redmon2016you,redmon2017yolo9000,Joseph2018Yolov3,bochkovskiy2020yolov4}.
Recently, anchor-free methods~\cite{law2018cornernet} and transformers~\cite{carion2020end} for object detection are widely used.

Multi-scale feature fusion is one of the most important research hotspots in deep networks. Low-level features generally lack semantic information but rich in keeping geometric details, which is the opposite for high-level features. FPN~\cite{lin2017feature} first built a top-down architecture with lateral connections to extract features across multiple layers. PANet~\cite{liu2018path} directly created a short path for low-level feature maps since detecting large objects also needs the assistance of location-sensitive feature maps. Libra R-CNN~\cite{pang2019libra} improved the level of feature fusion by adding a non-local block to fine-tune the combined feature maps.
\begin{figure}[h]
	\centering
	\includegraphics[height=0.4\textwidth]{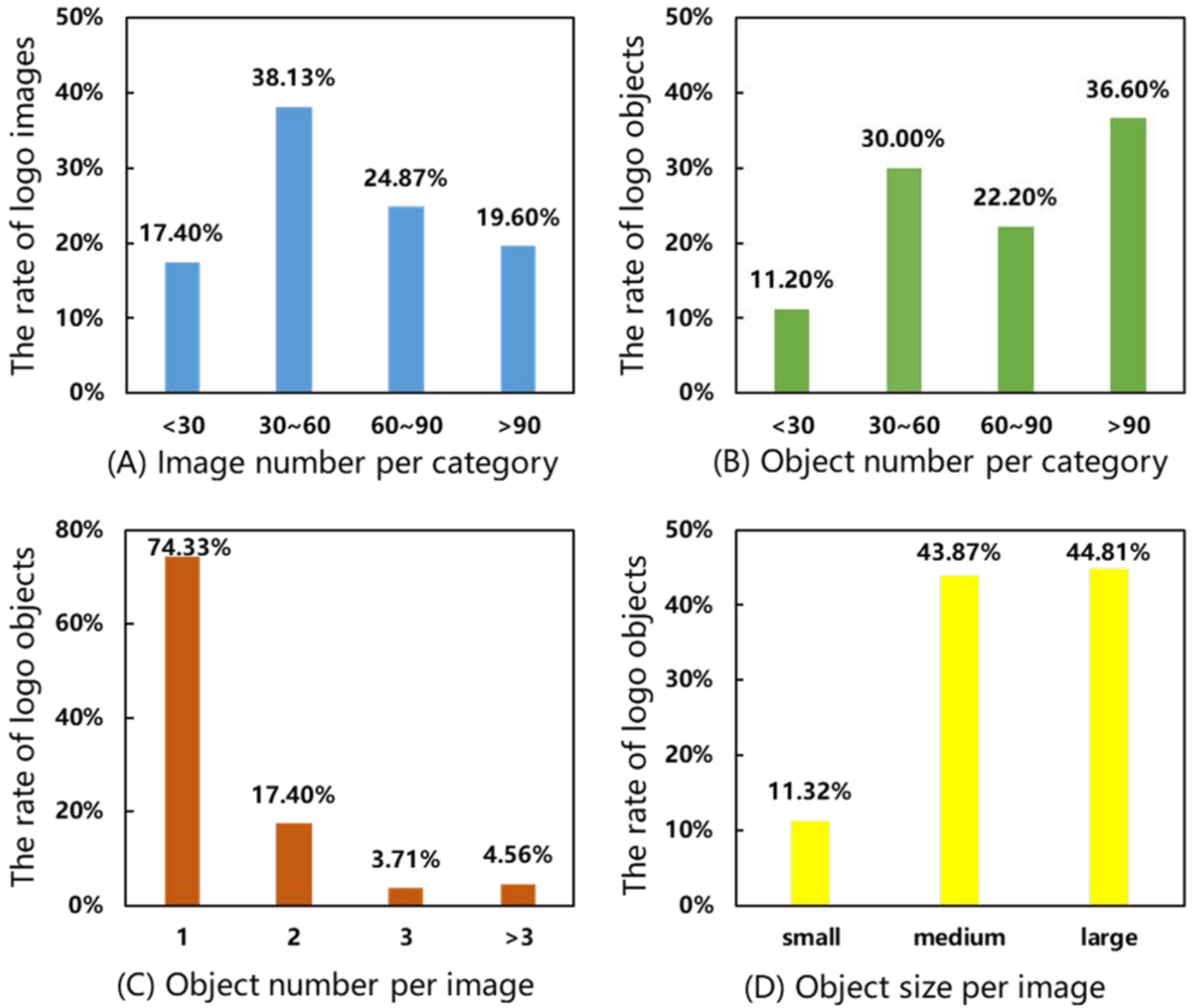}
	\caption{The detailed statistics of FoodLogoDet-1500.}
	\label{fenxi}
\end{figure}

Detection heads are also one of the focuses of research. Mask R-CNN~\cite{he2017mask} brought in an extra head for instance segmentation. IoU-Net~\cite{jiang2018acquisition} introduced a branch to predict IoUs between detected bounding boxes and their corresponding ground truth boxes. FCOS~\cite{tian2019fcos} added a single-layer branch, which is parallel to the classification branch, to predict the centerness position. Double-Head R-CNN~\cite{wu2020rethinking} proposed to disentangle the sibling head into two independent branches for classification and localization. Song~\textit{et al.}~\cite{song2020revisiting} also dealt with the classification branch and the regression branch and obtained a relatively good detection result.

Different from these works, our work decouples classification and regression into two branches and focuses on the classification branche to solve the problem of distinguishing multiple food logo categories. At the same time, we consider the multi-scale and similar characteristics of food logos, and use multi-layer features for food logo detection.

\section{FoodLogoDet-1500}
In order to obtain a high-quality food logo dataset with high diversity and high coverage. We build FoodLogoDet-1500 from the following three steps:~\textbf{(1) Constructing the Food Logo Category List.} In order to guarantee wide coverage of the food logo category list, we resort to the widely used shopping application Taobao and Jingdong, also with Wikipedia to construct the food logo category list.  ~\textbf{(2) Collecting Food Logo Images.} Using a query term from the constructed food logo category list, we crawled candidate images from various search engines ( i.e., Google, Bing and Baidu ) for broader coverage and higher diversity of food logo images compared with other datasets from only one data source. At the same time, we also added some scene words to ensure more complexity and better diversity of the captured food logo images, such as Coca Cola + Supermarket and Heineken + Bar.~\textbf{(3) Cleaning and Labeling Food Logo Images.} We checked each category manually to ensure that each image contained the corresponding food logo. It is worth noting that we focused on the food logo rather than the food itself. We also deleted repetitive images and images with incomplete RGB channels. Labeling is not only the most important step in creating a dataset, but also the most complicated step. Every food logo object needs to be annotated, regardless of which image it is placed on. We kept the low-resolution, incomplete food logo images to enhance the challenge of the dataset. After labeling, we then conduct manual verification by crowd-sourcing the task to 13 Lab members. In addition, a food brand may have two or more different types of logos, such as graphic logos and textual logos. We treat different logo variations of the same brand as distinct food logo classes, similar to~\cite{tuzko2017open}. Note that the suffix ‘-1’, ‘-2’ is added to the logo name as the new logo category, such as ‘Maruchan-1’ presents the ‘Maruchan’ graphic logo while ‘Maruchan-2’ presents its textual logo for the brand ‘Maruchan’.
\begin{figure*}
	\centering
	\includegraphics[width=0.85\textwidth]{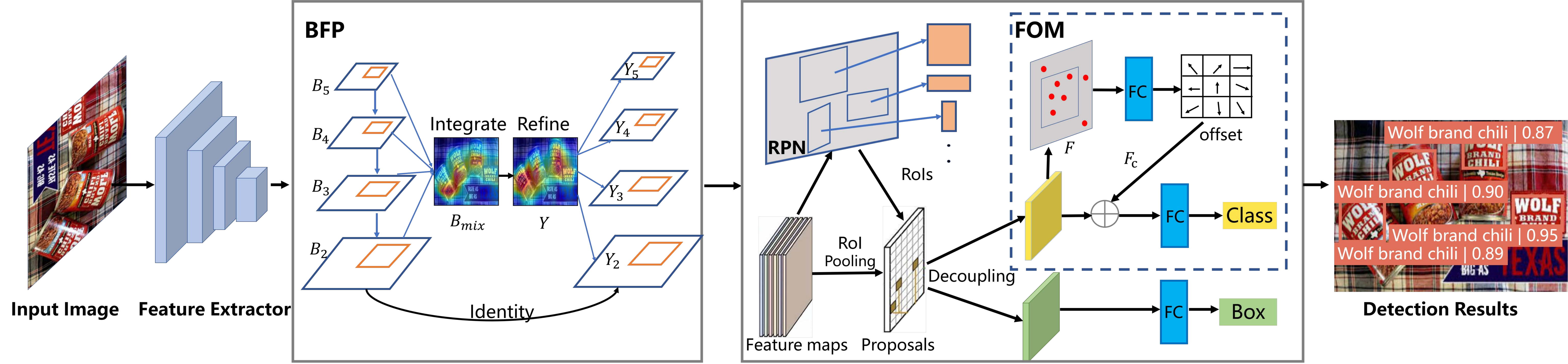}
	\caption{Overview of proposed Multi-scale Feature Decoupling Network (MFDNet) for food logo detection. BFP: Balanced Feature Pyramid. FOM: Feature Offset Module. RPN: Region Proposal Network. FC: Full-Connected layer. }
	\label{ds}
\end{figure*}

After completing the construction of the FoodLogoDet-1500, in order to show the details of our dataset, we provide the statistics at the category levels. Fig.~\ref{distribution} shows sorted distribution of the number of images from sampled classes, we can see that imbalanced distribution across different food logo categories are one characteristic of FoodLogoDet-1500, posing a challenge for effective food logo detection with few samples. In addition, we also conduct data statistics on images and objects in the FoodLogoDet-1500 as shown in Fig.~\ref{fenxi}. Fig.~\ref{fenxi} (A) shows the distribution of the number of images for each category, where each category represents each food logo. Fig.~\ref{fenxi} (B) shows the distribution of the number of objects for each category. As we can see, there is the imbalance between images and objects in different food logo categories. Fig.~\ref{fenxi} (C) provides the number of objects per image. We can draw the conclusion that most images contain one or two logo objects, which is similar to what happens in our real world. Fig.~\ref{fenxi} (D) gives the number of objects size in each image. In FoodLogoDet-1500, the large percentage of small and medium food logo objects ($\sim$~56\%) will pose another challenge to food logo detection, since smaller food logos are harder to detect.

\section{METHODOLOGY}
In this section, we will introduce the proposed Multi-scale Feature Decoupling  Network (MFDNet) for food logo detection. Fig.~\ref{ds} illustrates the architecture of MFDNet, which contains two main components, namely Balanced Feature Pyramid (BFP) and Feature Offset Module (FOM). Specifically, the features of one input food logo image are extracted by ResNet-50~\cite{he2016deep}. Then FPN is employed to fuse multi-scale features and BFP is used for feature refinement in the feature maps. Feature fusion and feature refinement are more effective for multi-scale food logo detection. The region proposal generation step yields a set of region of interests (RoIs) using Region Proposal Network (RPN). Then the RoIs are fed into RoI Pooling layer, in which each RoI is pooled into a fixed-size feature map. Finally, it is divided into classification and regression branches by feature decoupling. FOM is used to disentangle the classification and regression. In the classification branch, FOM utilizes the deformation-learning for optimal offset, which helps us to obtain the most representative features of classification in food logo detection. Then the optimal classification offset is merged with the original features of the network. Finally, feature maps are mapped to a feature vector by a fully connected layer (FC), which is followed by training the final object classifiers and bounding box regressors.

Next, we will focus on two main modules in the MFDNet, namely BFP and FOM.

\subsection{BFP}	 
In object detection, multi-scale features fusion has been a hot topic of research. Deep high-level features in backbones are with more semantic information while the shallow low-level features are more descriptive content ~\cite{zeiler2014visualizing}. On that basis, we fuse BFP into MFDNet to better integrate multi-scale features. Different from former methods that integrate multi-level features using lateral connections, the BFP uses the same deeply feature maps to integrate balanced semantic features to strengthen the multi-level features.

To integrate multi-level features while maintaining  their semantic hierarchy, we first adjust the multi-level features of FPN outputs $ \big\{B_2, B_3, B_4, B_5\big\} $ to the same size as $B_4$. This is achieved using interpolation and maxi-pooling on the other levels to prepare for integration. And then 
the integrated semantic information is obtained by Eq.~\ref{resize}.

\begin{equation}
	B_{mix}=\frac1L\;\sum_{l_{min}}^{l_{max}}B_l
	\label{resize}
\end{equation}
where $B_l$ is $l$-th feature maps. The number of multi-level features are denoted as $L$. $l_{min}$ and $l_{max}$ are the lowest and highest feature levels, respectively.

Then, the BFP uses non-local module to further refine balanced semantic features. The refining step helps us to enhance the integrated features which are more discriminative. The non-local module is adopted as follows:

\begin{equation}
	Y_i=\frac1{C(x)}\sum_{\forall j}f(B_{mix}^i,B_{mix}^j)g(B_{mix}^j)
\end{equation}
where $i$ is the index of an output position whose response is to be computed and $j$ is the index that enumerates all possible positions in feature map $B_{mix}$. $Y$ is the output of the same size as $B_{mix}$. $f(\cdot,\cdot)$ computes a scalar between $i$ and all $j$. $g(\cdot)$ computes a representation of the input at the position j. $C(x)$ is the normalization parameter.

After BFP, we can use the feature information of different layers more effectively.

\subsection{FOM}
The challenge of food logo detection lies in the large-scale categories and similar food logos. Thus, we focus on the problem of classification on food logo detection with large-scale categories. In food logo detection, we are more inclined to extract more expressive semantic regional features in images for large-scale food logo classification. As shown in Fig.~\ref{ds}, different from the original detection head, in FOM, we propose an auto-learned anchor region proposal network for pixel wise offset. FOM is used to search for the best feature extraction for food logo classification.

We used the deformable learning manner to achieve this goal. As shown in Fig.~\ref{ds}, $F$ is the output feature map of the RoI pooling layer. RoI pooling divides the RoI into 
$k\times k$ bins and output a $k\times k$ feature map $F_c$. From the $F$, a fully connected layer generates the normalized offsets $ \triangle\widehat{C}_{ij} $ which are then transformed to the offsets $ \triangle C_{ij} $ by element-wise product with the RoI’s width and height by Eq.~\ref{cl}. For $(i,j)$-th bin, the translation $\triangle C$ is performed on the sample points in it to obtain the new sample points for $F_c$. This procedure can be formulated as follows:
\begin{equation}
	\triangle C_{ij}=\alpha\triangle\widehat{C}_{ij}\;\cdot\;(w,h)
	\label{cl}
\end{equation}
where $\alpha$ is a predefined scalar to modulate the magnitude of the $ \triangle C_{ij}$, and $(w,h)$ is the width and height of $F$. 

For generating feature maps by irregular $F_c$, we use the deformable RoI pooling as:
\begin{equation}
	{F}_c(i,j)=\sum_{p\in bin(i,j)}\frac{G(p_{0\;},\;p_0+p_n+\triangle C_{ij})}{n_{ij}}
\end{equation}
where $p_0$ is the top-left corner and $p_n$ enumerates all integral spatial
locations in the feature map. $n_{ij}$ is the number of pixels in the bin.
As the offset $\triangle C_{ij} $ is typically fractional,
$G(\cdot,\cdot) $ is implemented via the bilinear interpolation.

By disentangling the shared proposal for the classification and regression, FOM is used to search for the best feature for food logo classification. It allows classification tasks to adaptively seek the optimal location in space. This has excellent detection accuracy for large-scale categories and similar food logos.

\subsection{Loss Function}
In MFDNet, the final loss function is as follows:
\begin{equation}
	L=L_{rpn}+L_{cls}+L_{loc}+L_{FOM}
	\label{loss}
\end{equation}
where $L_{rpn}$ , $L_{cls}$ and $L_{loc}$ are the losses for RPN, classification and localization, respectively. $ L_{FOM} $ is the loss for FOM.

Among them, the loss function of the FOM is as follows:
\begin{equation}
	L_{FOM} = L_{F}(C(f(F,F_c)),y)
\end{equation}
where $L_{F}$ is achieved through the cross-entropy loss function. $f(\cdot)$ is the feature extractor and $C(\cdot)$ is a function for transforming features to predict specific category. $y$ is the logo category.

The cross-entropy classification loss function is adopted as follows:
\begin{equation}
	L_{F}=-\frac1N\;\sum_i^M\;\log y_{ic}\left(p_{ic}\right)
\end{equation}
where $N$ is the number of training samples and $M$ is number of food logo categories. $y_{ic}$ is the indicator variable. If the sample category is the same as the category of sample $i$, then $y_{ic}$ is 1. $p_{ic} $ is the probability which predict the whether a sample belongs to category $c$.

\section{Experiment}
\subsection{Experimental Setup}
\begin{figure*}[htb]
	\includegraphics[width=0.8\textwidth]{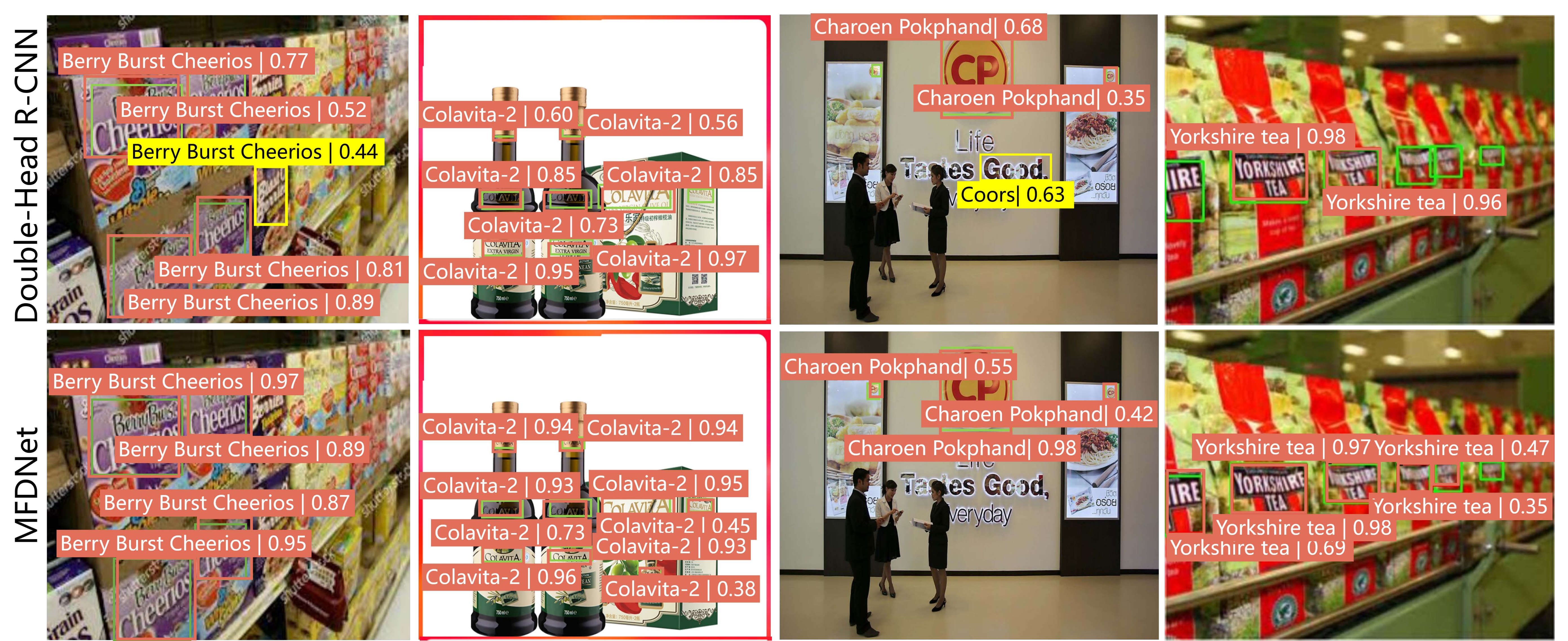}
	\caption{Visualization comparison between Double-Head R-CNN and MFDNet on the FoodLogoDet-1500. The first row is Double-Head R-CNN, the second row is MFDNet. Green boxes: ground-truth boxes. Orange boxes: correct detection boxes. Yellow boxes: mistakes detection boxes.}
	\label{duibi}
\end{figure*}
\textbf{Dataset and evaluation metrics.} 
To evaluate the effectiveness of the proposed MFDNet, we conduct extensive experiments on our introduced FoodLogoDet-1500 and two standard logo detection datasets, including the FlickrLogos-32 and the QMUL-OpenLogo. 

For evaluation, we adopt the widely used mean Average Precision (mAP)~\cite{Everingham2010voc} and the IoU threshold is 0.5, which means that 
a detection will be considered as positive if the IoU between the predicted
box and ground-truth box exceeds 50\%. We also use  AP$_{25} $ and  AP$_{75} $ as evaluation standards, which represent the IoU threshold of 0.25 and 0.75, respectively.

\textbf{Implementation details.} 
We implement our method based on the publicly available mmdetection toolbox~\cite{chen2019mmdetection}. The Double-Head R-CNN based on ResNet-50 is adopted as the baseline network.

In our experiment, the base detection networks are trained with stochastic gradient descent (SGD). The input images are resized to 800$ \,\times \, $600 pixels. 
We train detectors end-to-end with 2 GPUs (2 images per GPU) for 12 epochs. The initial learning rate is set to $2.5\times10^{-3}$. The weight decay of 0.0001 and the momentum of 0.9 are used. Other hyperparameters follow the settings in mmdetection unless otherwise specified.
\begin{table}[H]
	\caption{Evaluation on individual modules and two modules of
		MFDNet on FoodLogoDet-1500~(\%).}
	\setlength{\tabcolsep}{4mm}{
		\begin{tabular}{ccccc}
			\toprule
			FOM&BFP&mAP&AP$_{25}  $&AP$_{75} $     \\ \hline	
			&   & 84.5&84.4&82.1      \\
			\checkmark&  & 86.0$\uparrow_{1.5} $&85.9$\uparrow_{1.5}$&84.4$\uparrow_{2.3}$ \\
			& \checkmark & 85.1$ \uparrow_{0.6} $&85.0$ \uparrow_{0.6} $&83.1$ \uparrow_{1.0} $ \\
			\checkmark &\checkmark &\textbf{86.6$\uparrow_{2.1} $}&\textbf{86.4$\uparrow_{2.0} $}&\textbf{85.0$\uparrow_{2.9} $} \\
			\hline\\
	\end{tabular}}
	\label{ablation1}
\end{table}
\begin{table}[htbp]
	\caption{Performance comparison on FoodLogoDet-1500~(\%).}
	\setlength{\tabcolsep}{4mm}{
		\begin{tabular}{cccc}
			\hline
			Method  &mAP&AP$_{25}  $&AP$_{75} $\\ \hline
			Faster R-CNN~\cite{ren2015faster}&83.9&83.8&81.7\\
			RetinaNet~\cite{lin2017focal} &77.3&77.0&75.3\\
			DCN~\cite{dai2017deformable}&85.2&85.1&84.2\\
			Cascade R-CNN~\cite{cai2018cascade}&83.5&83.3&83.2\\
			PANet~\cite{liu2018path} &83.8&83.6&81.5\\
			Libra R-CNN~\cite{pang2019libra}&77.8&77.7&76.4\\
			FSAF~\cite{zhu2019feature}&83.0&83.5&81.0\\
			Dynamic R-CNN~\cite{zhang2020dynamic} &75.9&75.5&65.6\\
			Sparse R-CNN~\cite{sun2020sparse}&83.1&-&-\\
			SABL~\cite{wang2020side} &82.9&83.4&82.0 \\
			GRoIE~\cite{rossi2020novel}&83.4&83.6&82.1 \\
			Generalized Focal Loss~\cite{li2020generalized}&79.2&79.1&78.5\\
			Double-Head R-CNN~\cite{wu2020rethinking}&84.5&84.4&82.1\\
			ATSS~\cite{zhang2020bridging} &80.2&80.0&79.8 \\
			FoveaBox~\cite{kong2020foveabox}&75.2&75.0&74.0\\
			Soft-NMS~\cite{bodla2017soft}&83.8&83.9&81.8\\
			OHEM~\cite{shrivastava2016training}&84.1&84.2&82.5\\
			Iou loss~\cite{yu2016unitbox}&82.2&82.1&79.5\\
			Generalized IoU~\cite{rezatofighi2019generalized}&83.3&83.2&80.4\\
			SSD~\cite{liu2016ssd}&80.4&80.1&78.6\\
			\hline	
			\textbf{MFDNet}&\textbf{86.6}&\textbf{86.4} &\textbf{85.0}\\ \hline
			\label{1500}
	\end{tabular}}
\end{table}
\subsection{Experiment on FoodLogoDet-1500}
To enable the benchmark research, we follow the standard setup for data partitions in our experiments. 80\%, 20\% of images are randomly selected for training and testing in each food logo category. 

\textbf{Ablation Study.} For the ablation study, we conduct a comprehensive analysis of the effects of two modules from the MFDNet in the FoodLogoDet-1500. Table~\ref{ablation1} shows an ablation study on the effects of different combinations of FOM and BFP in the FoodLogoDet-1500. Two modules are added to Double-Head R-CNN, and the results respectively improve the mAP by 1.5\%, 0.6\% and 2.1\%.
These results prove the effectiveness of the FOM in large-scale food logo dataset.

Next, we perform the visualization of the ablation study and analyze the existing detail problem in the Double-Head R-CNN. And then we provide more typical examples in Fig.~\ref{duibi}, including the regression bounding box and the classification accuracy. The red box represents the prediction box and the green box is the ground-truth box. Clearly, MFDNet can accurately detect objects with occluded, ambiguous and smaller cases, and it obtains more accurate bounding box regression and classification score. The Double-Head R-CNN makes some detection mistakes, such as misclassified similar food logo categories, and mistaking words into food logo categories. Such as the word ‘Good’ is used as a food logo ‘coors’, because two words are similar. The reason for the above errors is that the large-scale categories of the food logo and the similarity between the food logos are not considered. In contrast, for the detected logos in the middle two images in Fig.~\ref{duibi}, our method has an advantage in classification accuracy, and also detect smaller food logos. This shows that FOM can search the best feature for classification, and BFP can fusion of multi-scale information for detection.

\begin{figure*}[htb]
	\includegraphics[width=0.8\textwidth]{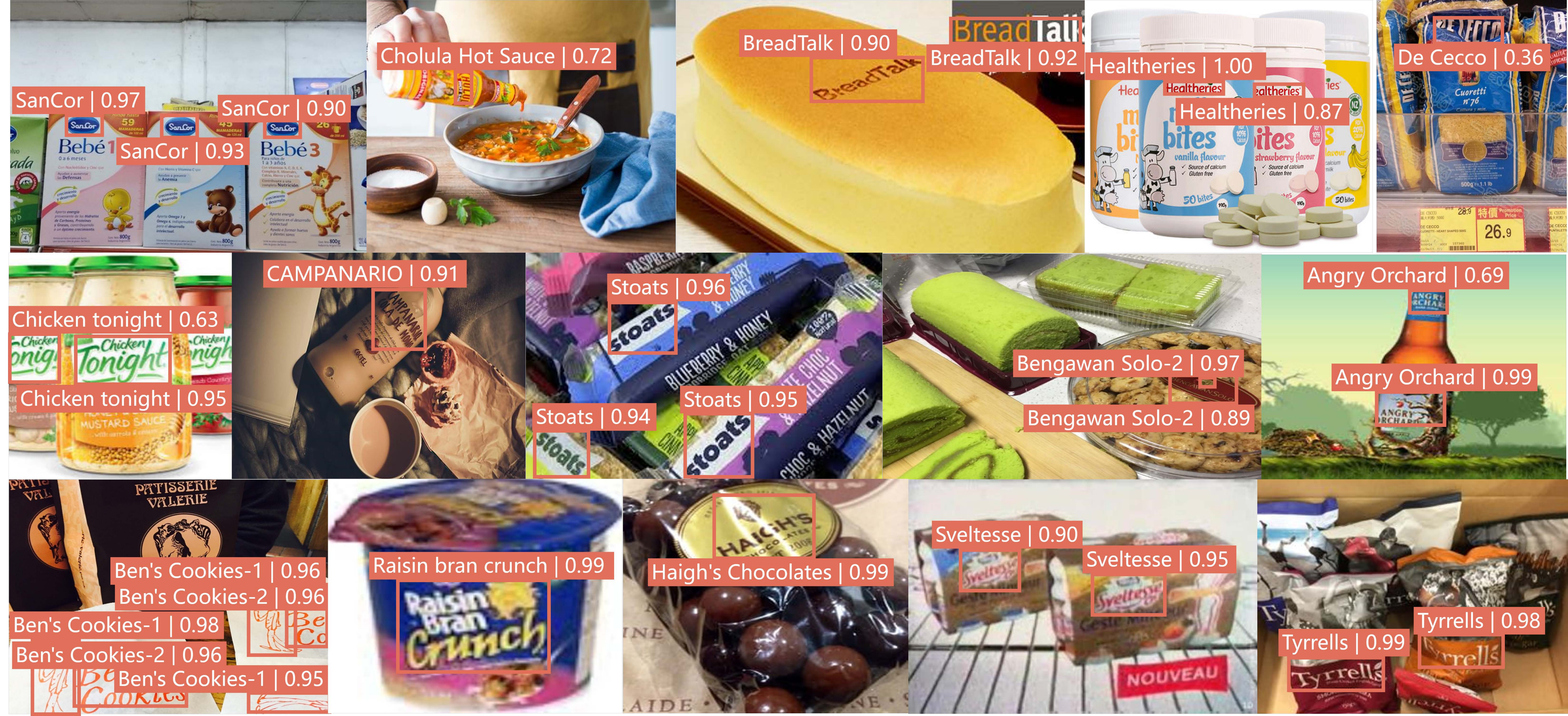}
	\caption{Detection results for our proposed MFDNet on the FoodLogoDet-1500. Orange boxes: correct detection boxes. }
	\label{demo}
\end{figure*}

\textbf{Comparisons with State-of-the-Arts.}
In this subsection, we compare the results of our method with other works in FoodLogo-Det-1500. Table~\ref{1500} summarizes the clear performance superiority of MFDNet over all state-of-the-arts with significant mAP, AP$_{25}$ and AP$_{75}$ improvement. SSD uses VGG-16~\cite{VGG2015} as a backbone, and other detection models adopt ResNet-50. Compared with existing baselines RetinaNet, Faster R-CNN and Double-Head R-CNN, etc., the proposed method significantly outperforms these state-of-the-art methods. RetinaNet had the limitation in multi-scale object detection. Faster R-CNN adopted the same parameters in two different tasks, which missed the conflict between them in the sibling head, especially the classification of large-scale datasets. MFDNet achieves the best performance with 86.6\% mAP. Compared with Double-Head R-CNN, our proposed method gains 2.1\% mAP. Specifically, it gains 2.0\% AP$_{25}$ and gains 2.9\% AP$_{75}$. This shows that our proposed method still has good performance with the change of the IoU threshold. MFDNet also surpasses methods of two-stage detectors (Faster R-CNN, Sparse R-CNN and GRoIE), boosting the mAP by 2.7\%, 3.5\% and 3.2\%, respectively. These results validate the advantage of our feature decoupling over existing methods. Some detection results of MFDNet are given in Fig.~\ref{demo}, including the regression bounding box and the classification accuracy. The red box represents the prediction box. 

\begin{figure}[htb]
	\includegraphics[width=0.4\textwidth]{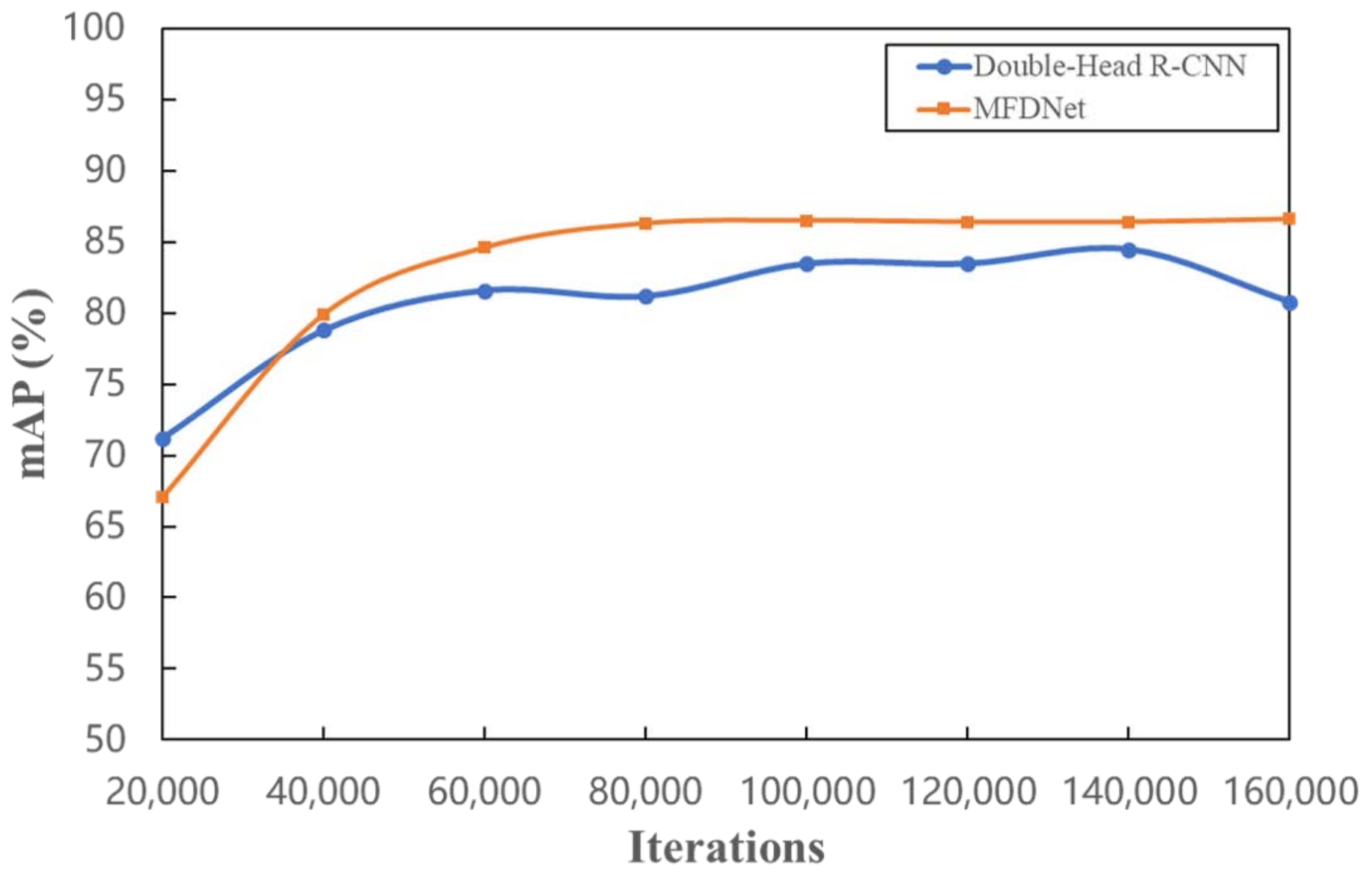}
	\caption{The comparison of MFDNet and Double-Head R-CNN with increasing iterations.}
	\label{map}
\end{figure}

We also set different iterations to compare the convergence and accuracy of models. Fig.~\ref{map} shows higher performance with increasing iterations. It can be seen that our method converges at about 100,000 iterations and keeps higher accuracy than Double-Head R-CNN in the training process. This shows that feature decoupling can speed up model convergence.

\subsection{Experiment on Other Benchmarks}
Besides FoodLogoDet-1500, we also conduct the evaluation on other publicly available benchmark datasets, QMUL-OpenLogo and FlickrLogo-32 to further verify the effectiveness of our method. QMUL-OpenLogo contains 27,083 images from 352 logo categories. In each logo category, 70\%, 30\% of images are randomly selected for training and testing, respectively~\cite{su2018open}.
FlickrLogos-32 consists of 2,240 images from 32 logo categories. 80\%, 20\% of images are randomly selected for training and testing for each logo category. Considering that these baseline experiments only used mAP as the evaluation metric, we also used mAP as the evaluation standard for comparison.

\textbf{Experiments on QMUL-OpenLogo.} We list the experimental results
of baselines and our proposed method in Table~\ref{open}. Our proposed method achieves the best performance with 51.3\% mAP. Specifically, MFDNet outperforms the baseline model by 0.4\% in mAP. Compared to the anchor-free method FSAF, our method improves the mAP by 6.6\%. These results demonstrate the universality of our method on the large-scale logo dataset. 
\begin{table}[H]
	\caption{Performance comparison on QMUL-OpenLogo~(\%).}
	\setlength{\tabcolsep}{4mm}{
		\begin{tabular}{cccc}
			\toprule
			Methods  & mAP\\ \hline
			YOLO9000~\cite{redmon2017yolo9000}&26.3\\
			ATSS~\cite{zhang2020bridging}  &48.4\\
			Faster R-CNN~\cite{ren2015faster} &51.2\\
			Libra R-CNN~\cite{pang2019libra} & 51.2\\
			FSAF~\cite{zhu2019feature}&44.7\\
			Dynamic R-CNN~\cite{zhang2020dynamic}& 51.2\\
			FoveaBox~\cite{kong2020foveabox}&35.6 \\
			Generalized Focal Loss~\cite{li2020generalized}&46.6 \\
			Sparse R-CNN~\cite{sun2020sparse}&46.9\\
			Double-Head R-CNN~\cite{wu2020rethinking} &50.9\\\hline
			\textbf{MFDNet}&\textbf{51.3}\\ \hline\\
	\end{tabular}}
	\label{open}
\end{table}

\textbf{Experiments on FlickrLogos-32.} To better prove the effectiveness of our method, we also carry out experiments on the FlickrLogos-32 with fewer images. Table~\ref{f32} shows that MFDNet still achieves the best performance compared with other methods, surpassing the Double-Head R-CNN by 0.9\% in mAP. However, our model achieves a small margin 0.1\% in mAP over the best method Generalized Focal Loss. The probable reason is that FlickrLogos-32 contains fewer logo images and the FOM module thus does not play a decisive role in the dataset.

\begin{table}[H]
	\caption{Performance comparison on FlickrLogos-32~(\%).}
	\setlength{\tabcolsep}{4mm}{
		\begin{tabular}{cccc}
			\toprule
			Methods  & mAP\\ \hline
			Bag of Words (BoW)~\cite{romberg2013bundle}&54.5\\
			Deep Logo~\cite{iandola2015deeplogo}&74.4\\
			BD-FRCN-M~\cite{oliveira2016automatic}&73.5\\
			YOLO~\cite{redmon2016you} &68.7\\
			YOLOv3~\cite{Joseph2018Yolov3}&71.7\\
			RetinaNet~\cite{lin2017focal}&78.4 \\
			Faster R-CNN~\cite{ren2015faster}&83.5\\
			Libra R-CNN~\cite{pang2019libra}&84.6\\
			Dynamic R-CNN~\cite{zhang2020dynamic}&85.8\\
			FoveaBox~\cite{kong2020foveabox}&84.1\\
			Generalized Focal Loss~\cite{li2020generalized}&86.2\\
			Sparse R-CNN~\cite{sun2020sparse}&73.7\\
			Double-Head R-CNN~\cite{wu2020rethinking}&85.3\\\hline
			\textbf{MFDNet}&\textbf{86.2}\\ \hline\\
	\end{tabular}}
	\label{f32}
\end{table}

\subsection{Discussion}
Compared with existing methods, our proposed method obtains better detection performance, especially in solving small food logo objects and large-scale classification. However, it can not achieve high detection performance in some cases. In Fig.~\ref{duibi}, the fourth image in the second row shows although the MFDNet improves the detection accuracy of small food logos, there are also missed detections for smaller objects. Therefore, the food logo detection on FoodLogoDet-1500 still has great challenges, such as the problem of the small food logo objects. And it meanwhile highlights the comparative difficulty of the FoodLogoDet-1500.

\section{Conclusions}
In this paper, we present a new large-scale dataset FoodLogoDet-1500, which 
is currently the first and largest publicly available food logo detection dataset to the best of our knowledge. In the future, we hope FoodLogoDet-1500 will become a new benchmark food logo dataset, and provide convenience for food logo detection. We then propose a Multi-scale Feature Decoupling Network for food logo detection. Extensive evaluation on FoodLogoDet-1500 and another two standard benchmark logo datasets have verified its effectiveness.

With the rapid development of e-commerce platforms and major food brands, food logo detection will become the trend of future research. We will continue to explore the characteristics of the FoodLogoDet-1500, and generate different benchmarks to evaluate its challenges, such as tiny food logo, serious occlusion and low resolution. Furthermore, we will use transformer~\cite{carion2020end} and lightweight methods to achieve faster and more accurate  performance for food logo detection.

\begin{acks}
	This work was supported in part by the National Nature Science Foundation of China (62072289, 61702313, and 61972378), in part by Postdoctoral Science Foundation of China (2017M612338), in part by Shandong science and technology plan project (J17KB177).
\end{acks}

\bibliographystyle{ACM-Reference-Format}
\balance
\bibliography{sample-base}

\end{document}